\documentclass[twoside,11pt]{article}
\usepackage{jmlr2e_no_editor}
\usepackage[utf8]{inputenc} 
\usepackage[T1]{fontenc}    
\usepackage{hyperref}       
\usepackage{url}            
\usepackage{booktabs}       
\usepackage{nicefrac}       
\usepackage{graphicx} 
\usepackage{subfigure} 
\usepackage{natbib}
\usepackage{amsmath}
\usepackage{bm}
\usepackage{amsfonts}
\usepackage{amssymb}
\usepackage{caption}
\DeclareMathOperator{\Tr}{Tr}
\let\oldhref\href
\renewcommand{\href}[2]{\oldhref{#1}{\hbox{#2}}}
\usepackage{algpseudocode,algorithm,algorithmicx}
\algdef{SE}[DOWHILE]{Do}{doWhile}{\algorithmicdo}[1]{\algorithmicwhile\ #1}%

\let\oldhref\href
\renewcommand{\href}[2]{\oldhref{#1}{\hbox{#2}}}

\ShortHeadings{DP Regression and Classification with Sparse GPs}{Smith, \'{A}lvarez and Lawrence}
\firstpageno{1}

\begin{document}
\title{Differentially Private Regression and Classification with Sparse Gaussian Processes}

\author{\name Michael Thomas Smith${}^1$ \email m.t.smith@sheffield.ac.uk \\
       \name Mauricio A. \'{A}lvarez \email mauricio.alvarez@sheffield.ac.uk \\
       \name Neil D. Lawrence${}^2$ \email neil@sheffield.ac.uk \\
       \addr Department of Computer Science\\
       University of Sheffield, UK}


\maketitle

\begin{abstract}
A continuing challenge for machine learning is providing methods to perform computation on data while ensuring the data remains private. In this paper we build on the provable privacy guarantees of differential privacy which has been combined with Gaussian processes through the previously published \emph{cloaking method}. In this paper we solve several shortcomings of this method, starting with the problem of predictions in regions with low data density. We experiment with the use of inducing points to provide a sparse approximation and show that these can provide robust differential privacy in outlier areas and at higher dimensions. We then look at classification, and modify the Laplace approximation approach to provide differentially private predictions. We then combine this with the sparse approximation and demonstrate the capability to perform classification in high dimensions. We finally explore the issue of hyperparameter selection and develop a method for their private selection. This paper and associated libraries provide a robust toolkit for combining differential privacy and GPs in a practical manner.
\end{abstract}


\section{Introduction}
{\footnotetext[1]{Corresponding author.}}
{\footnotetext[2]{Work conducted while at the University of Sheffield.}}

Simple `anonymisation' (e.g. removing names and addresses) has been found to be insufficient for protecting the privacy of individuals when releasing statistics from a dataset \citep{sweeney1997weaving, ganta2008composition}. Instead one should consider randomisation-based privacy methods (such as differential privacy, DP) which provide provable protection against such attacks.

In a previous paper \citep{smith18}, the authors introduced a method for differentially private Gaussian process (GP) regression, called the \emph{cloaking mechanism}, in which they corrupted a Gaussian processes's posterior mean in order to make aspects of the training data private. Specifically this method protects the \emph{outputs} of the training data. An example application might be inference over census data. The existence of a property at a location, $\bm{x_i}$, is not private. However a feature, $y_i$, about the household who live there is. Hence the training inputs, $X$, to the algorithm are public while the training outputs, $\bm{y}$, are private. Another example is the results of a school's mental health assessment each year, to investigate if the numbers are increasing. The school year a child is in is non-private, but their mental health state is. 

The previous paper identified several shortcomings or areas for future work. Specifically the method was susceptible to outliers and produced predictions with unsatisfactory levels of noise added to provide DP. Second, the method was only applicable to the Gaussian likelihood problem of regression, and not to the issue of classification. Third, although briefly discussed in the supplementary, hyperparameter optimisation was not fully handled.

In the current paper we suggest solutions to some of that method's shortcomings, specifically we propose non-stationary covariance functions to reduce the impact of outliers, and develop a method for using the Laplace approximation to perform classification. We describe a method for performing hyperparameter optimisation and finally combine a sparse approximation with the classification result to allow the algorithm to function on a high dimensional classification problem.

\subsection{Related Work}
\label{relatedwork}
It is largely assumed that regression `involves solving an optimization problem' \citep{zhang2012functional} and as discussed previously \citep{smith18} there is little work currently proposing methods for DP non-parametric regression. We believe that, by avoiding a parametric formulation, we can also largely avoid the problem of iterative optimisation which the majority of parametric methods require. Similarly most examples for DP classification  \citep[e.g.][]{chaudhuri2011differentially, rubinstein2009learning, song2013stochastic}, work by perturbing the objective function and constraining the DP noise scale by carefully crafting the loss-function. In this paper we build on the cloaking method from \citet{smith18}, a non-parametric (GP based) method for DP regression, as an alternative route to perform DP machine learning. Note that all the papers \citep[besides][]{smith18} consider the problem of making both the inputs and outputs of the training data private, thus aren't directly comparable to the cloaking method or its improvements outlied in this paper, which only consider output privacy.

The most common non-parametric DP regression method is the histogram, used as a baseline in \citet{smith18} and is described in \citet{lei2011differentially}. In both of these papers a non-DP regression algorithm is fitted to DP corrupted results of the histogram. When compared, the cloaking method of \citet{smith18} required less noise to achieve DP than using this intermediate histogram, although the histogram method is easy to extend to include privacy on the inputs, something absent in the cloaking method. Both the histogram and cloaking method are limited to low-dimensional input domains; the histogram method due to the exponential number of bins that are required as dimensionality grows (and associated low-occupancy); the cloaking method due to excessive sensitivity to outliers. This paper proposes an alteration to the cloaking method to allow higher-dimensional training data.

DP non-parametric \emph{classification} also has relatively little work associated. Although not truly non-parametric, \citet{li2014privacy} develop a protocol for computing the SVM's (linear) decision boundary normal vector in a DP manner. DP collaborative-filtering kNN methods have similarities and are non-parametric, and can be considered to be solving a classification problem. Examples include \citet{zhu2014effective} and \citet{afsharinejad2018performance}. In the former, the selection of neighbours is made DP using the exponential mechanism. In the latter the training points  are perturbed first (by corrupt encoding them) before performing a nearest neighbour operation. Then, in both, additional perturbation is conducted to protect those neighbours selected. DP parametric classification methods are more common. 
Unlike this paper's aims, the work in this field has focused on classifiers without strong priors, for example using deep neural networks \citep{abadi2016deep}, decision trees \citep{jagannathan2009practical, fletcher2017differentially}, or using a DP Bayes classifier \citep{vaidya2013differentially}. These and other results, e.g. \citet{chaudhuri2011differentially}, provide algorithms which ensure the training \emph{inputs} remain private too. Both of these aspects (input privacy and weak priors) mean that larger numbers of training points have been required. For example \citet{abadi2016deep} used 50,000-60,000 training points; \citet{chaudhuri2011differentially} used 47,000 and 5 million; \citet{zhang2012functional} used 190,000 and 370,000 (with linear and logistic regression); \citet{afsharinejad2018performance} used over 100,000 (kNN). 

So to emphasise, this paper addresses questions and problems of a fundamentally different nature. First, we are interested in the problem of privacy on just the \emph{outputs}. Second, our method targets the problem of learning from small datasets, unlike the above cited papers, in all our examples N<300.

\subsubsection*{Gaussian Processes}

A Gaussian process is a collection of random variables in which any finite subset have a Gaussian distribution. One can use this framework to perform regression by making assumptions about this distribution (for example that variables associated with nearby points are highly correlated). Then conditioning on observations for a portion of the subset can make predictions for the remainder.
 
In GP regression the covariance between predictions ($\bm{y}_*$) at test points only depend on the locations of the inputs of the training and test data $X$ and $X_*$. The mean however also depends on the training data's output values $\bm{y}$. Specifically, $\bm{y}_* = K_{*f} K^{-1} \bm{y}$ where $K_{*f}$ are the covariances between test and training inputs while $K$ is the covariance between training points. In this paper we refer to the product of these two matrices as the \emph{cloaking matrix}, $C = K_{*f} K^{-1}$, which describes how the test points change with respect to changes in training data.

One aspect that has been overlooked in the above description is how the observations are linked to the underlying latent function. In the example above we might assume that observations are points on a latent function, perturbed by Gaussian observation noise. Conveniently this leads to a posterior distribution that remains Gaussian. However, if the output represented (for example) the class of an observation then the distribution between random variables would no longer be Gaussian. We handle this by assuming that the latent function is `squashed' through a link function to give us the probability of the observation's classes. However, to then perform inference we need to approximate this posterior. We will discuss a particular method (the Laplace approximation) and a DP modification in Section \ref{GPClassification}.

\subsubsection*{Differential Privacy}
Differential privacy is defined as follows. From \citet{dwork2014algorithmic}, to query a database in a DP manner, a randomised algorithm $R$ is $(\varepsilon, \delta)$-differentially private if, for all possible query outputs $m$ and for all neighbouring databases $D$ and $D^\prime$ (those databases which only differ by one individual or one row),
$$
P\Big( R(D) \in m \Big) \leq e^{\varepsilon} P\Big( R(D^\prime) \in m \Big) + \delta. \label{dpdef}
$$
This simply states that we require each possible output to have almost the same probability regardless of the value of one row or individual. $\varepsilon$ puts a bound on how much privacy is lost by releasing the result, with a smaller $\varepsilon$ meaning more privacy. $\delta$ says this inequality only holds with probability $1-\delta$.

\subsubsection*{The Cloaking Method}
\label{thecloakingmethod}
\citet{hall2013differential} extended DP to apply to vectors. Given a covariance matrix $M$ which describes the Gaussian we will sample our (scaled) DP perturbation from, and result vectors describing neighbouring databases, $\bm{y}_*$ and $\bm{y}_{*}'$, we define the bound,

\begin{equation}
\sup_{D \sim {D'}} \Vert M^{-1/2} (\mathbf{y}_* - \mathbf{y}_{*}')\Vert_2 \leq \Delta.
\label{DPdelta}
\end{equation}

$\Delta$ is a bound on the scale of the output change, in term of its Mahalanobis distance with respect to the added noise covariance. The algorithm outputs
$
\tilde{\mathbf{y}}_* = \mathbf{y}_* + \frac{\text{c}(\delta)\Delta}{\varepsilon}Z
$
where $Z \sim \mathcal{N}_d(0,M)$. $(\varepsilon, \delta)$-DP is achieved if $\text{c}(\delta) \geq \sqrt{2 \log \frac{2}{\delta}}$.

We want $M$ to be as small as possible but still have covariance in those directions most affected by changes in training points, as that will allow us to have a small $\Delta$.

\citet{smith18} built on this bound to produce \emph{cloaking}; a method for making the training outputs of a GP regression calculation differentially private. We assume one training item $i$ has been perturbed, by at most $d$; $y_i' = y_i + d$. The change in the predictions is dependent on only one column of $C$, $\mathbf{c}_i$;
$
\mathbf{y}_*' - \mathbf{y}_* = d \mathbf{c}_i
$
This can be substituted into the bound on $\Delta$ in \eqref{DPdelta}. Rearranging the expression for the norm (and using $M$'s symmetry);
\vspace{-2mm}
$$
\Vert d M^{-1/2} \mathbf{c}_i \Vert_2 = d^2 \mathbf{c}_i^\top M^{-1} \mathbf{c}_i
\vspace{-2mm}
$$

We want to find $M$ such that the noise sample Z is small but also that $\Delta$ is also minimised. We describe the noise scale using the log of the determinant $\frac12 \ln\left((2\pi e)^k \cdot\left|\boldsymbol\Sigma \right|\right)$. It was shown in \cite{smith18} that the optimal $M = \sum_i{\lambda_i \mathbf{c}_i \mathbf{c}_i^\top}$, where $\bm{\lambda}$ is found using gradient descent,
\begin{equation}
\label{graddesc}\frac{\partial {M^{-1}}}{\partial \lambda_j} = - \Tr \left( M^{-1} \bm{c}_j \bm{c}_j^\top \right) + 1.
\end{equation}

\subsection{Paper Structure}

The paper is organised as follows: In Sections \ref{nonstat_kerns}-\ref{hypersel} we introduce the three topics, Nonstationary covariance, classification and parameter selection respectively. In Section \ref{results} we describe a series of experiments. Specifically Sections \ref{kungdemo} and \ref{kungdemo2} investigate using nonstationary covariances for reducing DP noise, while Section \ref{notesonvar} looks in depth at why variable lengthscales fail to achieve the improvements expected. Sections \ref{classification1} and \ref{classification2} look at DP GP classification. Section \ref{mnist} combines sparse approximation with classification. Finally Section \ref{hyperparameter} demonstrates how DP parameter selection might be performed.

\section{Nonstationary Kernels}
\label{nonstat_kerns}
To quickly reiterate the problem. We have a set of training data, consisting of public inputs, $X$, and private outputs, $\bm{y}$. We wish to make a prediction for a hypothesised latent function at location $\bm{x}_*$, using this training data, in such a way that an attacker will not gain much information about an individual value in $\bm{y}$.

In Sections \ref{kungdemo} and \ref{kungdemo2} we compare this paper's methods with standard cloaking (figures \ref{cloaking}A and \ref{cloaking_figures}A). These illustrate how, using the standard cloaking method from \citet{smith18}, DP noise becomes unreasonably large in the regions \emph{neighbouring} the data concentrations. This is due to the pivoting effect the data concentrations have on the outliers. Figure \ref{cloaking_figures}A demonstrates how this problem worsens as the number of dimensions increases. With additional dimensions each data point has more opportunities to become an outlier. In such areas of the domain only a small number of training points contribute towards the prediction (for a local kernel). Far more noise is then required to protect the individual training points.

We propose that this effect can be somewhat mitigated by manipulating the aggregation performed by the Gaussian process. Consider the structure of $C$, this simply describes the amount each training point contributes to a prediction. If we imagine there are $n$ outlying training points close together next to our test point then we will discover that their associated columns in $C$ will all be approximately equal to $1/n$. Effectively they will average the training outputs. If more training data is within this clique then the contribution provided by each training point is reduced. Obviously real data will lead to more complicated covariance behaviour, but a similar effect occurs - if more data lies within one-lengthscale of test point, then the effect of individual training points is diminished.

We initially consider using a non-stationary lengthscale to maintain approximately similar numbers of training points in the aggregation, effectively controlling for the density of training data at the test point location. We look at why this approach fails to achieve the desired aggregation.

As an alternative we consider the use of a sparse approximation using inducing inputs, placed over regions of high training density, through which the predictions are mediated. This has the practical effect of diminishing the contribution from those points far from the inducing inputs (i.e. the outliers).

Both of these methods can be considered a form of non-stationary kernel, as they both manipulate the influence of the training points in a way that varies depending on location.

\subsection{Non-stationary lengthscale}

One can manipulate the weighting of training data by using a non-stationary lengthscale, adjusted depending on the density of the data. 
Here we are interested in using lengthscales that are effectively fixed by the training data's input locations. \citet{paciorek2004nonstationary} describe a method, based on \citet[p48]{gibbs1998bayesian}, to produce a positive definite covariance matrix, if one already has \emph{a priori} beliefs about the lengthscale at each location. They propose that one considers the covariance of each of the locations $\bm{x}_i$ and $\bm{x}_j$ with a third point $\bm{u}$, over which we integrate;
$$
C(\bm{x}_i,\bm{x}_j) = \int_{\mathbb{R}^d} k(\bm{x}_i,\bm{u}) k(\bm{x}_j,\bm{u}) d\bm{u},
$$
where $k(\bm{a},\bm{b})$ is the covariance function, centred at $\bm{a}$.
We then need to select a lengthscale function, which gives the lengthscale for a given location; $l(\bm{x})$. If we knew the density, $\rho(\bm{x})$, of data at location $\bm{x}$ then we would want the lengthscale to be roughly $l(\bm{x}) = n / \rho(\bm{x})$, where $n$ is the number of training points we want within half a lengthscale (in 1d) of $\bm{x}$. To avoid arbitrarily large lengthscales, we rearrange and add a small constant $m^{-1}$ to the denominator, where $m$ then is an upper bound on the lengthscale; $l(\bm{x}) = [m^{-1} + \rho(\bm{x})/n]^{-1}$. Finally, we approximate $\rho(\bm{x})$ using KDE and an exponentiated quadratic (EQ) kernel. We found this initial design functioned poorly as long lengthscale areas were effectively detached from the high-density short lengthscale areas. In an attempt to ameliorate this we modified the function further during experimentation (see Section \ref{kungdemo}) by using the smallest value of $\rho(\bm{x})$ from a region neighbouring $\bm{x}$ in an attempt to allow higher density regions to support the low-density areas.

Section \ref{notesonvar} considers, as an alternative, a weighted sum of GPs as proposed by \citet{herlands2016scalable} in which the proportion of each function varies over the domain.

\subsection{Sparse Gaussian Processes}

An alternative way to manipulate the kernel over the input space is through the use of a sparse approximation. By using inducing inputs we effectively down-weight training data that lies far from these inputs. Thus, by simply placing the inducing inputs at locations of high density, we are able to reduce the influence of the outliers, and thus reduce the noise required for privacy. This method allows us to continue to use our chosen covariance function and parameters. However it does require selecting the number and location of inducing inputs, and it introduces biases towards training data that lies at the most dense locations.

To summarise, the effect of outliers can be diminished by pushing the regression through a sparse approximation, for example the Fully Independent Training Conditional (FITC) Approximation \citep{snelson2006sparse}.\footnote[2]{The model remains a GP (with the same covariance function for training and test) if we ignore the covariance between test points.} The GP's predictive distribution is now a normal with mean and variance,
\begin{equation}
\mu_* = \bm{k}_*^\top Q_{MM}^{-1} K_{MN} (\Lambda + \sigma^2 I)^{-1} \bm{y}
\label{meanfn}
\end{equation}
$$
\sigma^2_* = K_{**} - \bm{k}_*^\top (K_{MM}^{-1} - Q_{MM}^{-1}) \bm{k}_* + \sigma^2.
$$
Where, $Q_{MM} = K_{MM} + K_{MN} (\Lambda + \sigma^2 I)^{-1} K_{NM},$
$\Lambda = \text{diag}(\bm{\lambda})$
and
$\lambda_n = K_{nn}-\bm{k}_n^\top K_{MM}^{-1} \bm{k}_n.$
The $K_{NM}$ covariances are between training and pseudo inputs, while $K_{MM}$ are between the pseudo inputs, $\bm{k}_{*m}$ is between the test and pseudo inputs and $\bm{k}_n$ is between the training inputs and training point $n$. $K_{nn}$ is the variance of training point $n$.

As before, the variance term only depends on the inputs. Thus we only need to protect the expression for the mean. The cloaking matrix is now all the terms to the left of $\bm{y}$ in \eqref{meanfn},
\begin{equation}
C = \bm{k}_{*m}^\top Q_{MM}^{-1} K_{MN} (\Lambda + \sigma^2 I)^{-1}.
\end{equation}
We can again use the cloaking method, described in section \ref{relatedwork} to find the optimal $M$, using gradient descent, using \eqref{graddesc}.

Later, in Section \ref{kungdemo}, we will see the GP regression results with five inducing inputs placed using k-means clustering. The intuition is that these will then be placed sufficiently far from outliers that it is just the densest regions which will have much influence on the mean.

\section{Binary GP Classification}
\label{GPClassification}

We next investigated methods to apply DP to GP classification. In GP classification one assumes that a latent function $f(\bm{x})$ exists which is `squashed' through a link function to give us a distribution over the class probabilities $\pi(\bm{x})$. However this leads to an analytically intractable integral over the likelihood. This is therefore approximated, either numerically, or using analytical approximations, such as the Laplace approximation method or expectation propagation. To incorporate DP we chose to use the Laplace approximation, as it requires very few iterations for convergence and has a clear and simple relationship between the training outputs and the update step's direction, which is also of a form amenable to the application of the cloaking method.

From \citet{williams2006gaussian}, to apply the Laplace approximation we replace the exact posterior distribution $p(\bm{f}|X,\bm{y})$ with an approximation $q(\bm{f}|X,\bm{y}) = \mathcal{N}(\bm{f}|\bm{\hat{f}}, A^{-1})$, where $\bm{\hat{f}}$ is the mode of $p$, and $A$ is the Hessian of the negative log posterior at the mode. There is no analytical expression to find the mode. Instead, an algorithm is used in which an initial estimate for $\bm{\hat{f}}$ is iteratively updated,
\begin{equation}
\bm{\hat{f}}^{new} = (K^{-1} + W)^{-1} (W\bm{\hat{f}} + \nabla \log p(\bm{y}|\bm{\hat{f}})),
\label{fhatequation}
\end{equation}

where $K$ is the covariance between training input points and $W$ is a diagonal matrix; $W = -\nabla \nabla \log p(\bm{y}|\bm{f})$. The elements of $W$ for the logistic link function equal $-\pi_i(1-\pi_i)$ where $[\pi(\bm{\hat{f}})]_i = \pi_i = p(y_i = 1 | f_i) = (1+e^{-f_i})^{-1}$. To make this update differentially private we need to consider how the training outputs $\bm{y}$ alter $\bm{\hat{f}}^{new}$. If we are using the logistic link function, the gradient term $[\nabla \log p(\bm{y}|\bm{\hat{f}}))]_i = t_i - \pi_i$, where $\bm{t} = \bm{y}/2 + \bm{1}/2$, so we can replace occurrences of $\nabla \log p(\bm{y}|\bm{\hat{f}}))$ with $\bm{y}/2 + \bm{1}/2 - \pi(\bm{\hat{f}})$.

We define 
\begin{equation}
C = \frac{1}{2}(K^{-1} + W)^{-1}.
\label{Cdef}
\end{equation}
Substituting in this definition and the gradient term replacement into \eqref{fhatequation} we can rewrite the update as,
$$
\bm{\hat{f}}^{new} = 2C (W\bm{\hat{f}} + \bm{y}/2 + \bm{1}/2 - \pi(\bm{\hat{f}})) = 2C (W\bm{\hat{f}} + \bm{1}/2 - \pi(\bm{\hat{f}})) + C\bm{y}.
$$
Here the last term is the only one that depends on $\bm{y}$ (both $K$ and $W$ are independent of $\bm{y}$). Thus we only need to consider the effect of changes in $\bm{y}$ on $C\bm{y}$. We can use the \emph{cloaking method}, described in Section \ref{thecloakingmethod}, with a cloaking matrix as defined in \eqref{Cdef}. This allows us to estimate $\bm{\hat{f}}$ in a DP manner.

To make a prediction we need to compute the mean and variance of the latent function at a test point $\bm{x}_*$. We compute $\bm{k_*}$, the covariances between this test point and the training inputs. In \citet{williams2006gaussian} the value of the latent function's mean at a test point is computed as $\mathbb{E}_q[f_*] = \bm{k_*}^\top \nabla \log p(\bm{y}|\bm{\hat{f}})$. However, the gradient term depends on the values of $\bm{y}$.  To avoid this additional leakage of privacy, we instead use the approximation $\nabla \log p(\bm{y}|\bm{\hat{f}}) = K^{-1} \bm{\hat{f}}$ \citep[][eq 3.21. At the mode the gradient is zero and so eq. 3.13 is zero]{williams2006gaussian}, for which we have already used our privacy budget. So the mean of the posterior $\mathbb{E}_q[f_*] = \bm{k}_*^\top K^{-1} \bm{\hat{f}}$, and the variance $\mathbb{V}_q[f_*] = k(\bm{x}_*,\bm{x}_*) - \bm{k}_*^\top (K+W^{-1})^{-1} \bm{k}_*$. 
Entries in $\bm{\hat{f}}$, associated with nearby training points, will have similar values due to the covariance structure in $C$ and so even if there are quite large positive and negative values in $K$'s inverse, they shouldn't overwhelm the prediction. As explained in Section \ref{thecloakingmethod}, if there are more training points in a particular area around a point $i$, the DP cloaking approximation to $[C\bm{y}]_i$ for that element will be more accurate than for elements associated with outliers.

Note that for numerical stability, we use several alternative but equivalent expressions, as recommended in \citet[][pages 45-47]{williams2006gaussian}. Specifically (defining $B = K + W^{1/2} K W^{1/2}$) we compute $C = \frac{1}{2} (K - K (W^{1/2} B^{-1} W^{1/2} K))$. See reference for additional tricks to achieve numerical stability and computational efficiency.

\subsection{Sparse GP Classification}
We can modify the above method to incorporate a sparse approximation similar to the one introduced earlier. We simply substitute a low-rank approximation for $K$, following either the Subset of Regressors (SoR) approximation or the Deterministic Training Conditional (DTC) approximation \citep{quinonero2005unifying, seeger2003fast};

\begin{equation}
K = K_{NN'} K_{N'N'}^{-1} K_{NN'}^\top.
\end{equation}

Where $K_{NN'}$ is the covariance between the $N$ training inputs and $N'$ pseudo inputs; $K_{N'N'}$ is the covariance between pseudo inputs.\citet{quinonero2005unifying} note that the SoR prior can cause degeneracy in the prediction of the posterior variance, but we are only interested in the posterior \emph{mean} for the purposes of this component, as it is only the mean that is affected by the value of the training data's outputs. We place the $N'$ inducing points using k-means clustering, as above, as this only depends on the training inputs (which are not private in this scenario). The posterior covariance can also be computed using one of the sparse methods, if required, in a non-DP manner.

\section{Hyperparameter Selection}
\label{hypersel}
So far in this paper we have selected the values of the kernel hyperparameters \emph{a priori}. Normally one may maximise the marginal likelihood to select these values or potentially integrate over the hyperparameters. In differential privacy we must take care when using private data to make these choices. Previous work exists to perform this selection, for example \citet{kusner2015differentially} describes a method for performing differentially private Bayesian optimisation, however their method assumes the training data is not private.
\citet{kusner2015differentially} do suggest that the work of \citet{chaudhuri2013stability} may allow Bayesian optimisation to work in the situation in which the training data also needs to be private, we outline here a more direct solution.

\citet{chaudhuri2011differentially} proposed a tuning algorithm for selecting the classifier's (hyper)parameters by using the exponential mechanism. This maximises the probability of selecting the optimum hyperparameters while still ensuring DP. With modification this could apply to selecting hyperparameters for \emph{regression}. The bounding of the sensitivity requires a little more thought than in the classification case though. 

Due to the low-dimensionality of many hyperparameter problems, a simple grid search, combined with the exponential mechanism may allow the selection of an acceptable set of hyperparameters. To summarise, from \citet{dwork2014algorithmic}, the exponential mechanism is a method for selecting in a DP way the optimal (highest) choice from a set of utilities, in such a way that the values being selected over remain private. One must provide the sensitivity of the utility (i.e. an upper bound on how much a training point can change the utility). Using the utilities and their sensitivities, the exponential mechanism selects each item with a given probability, such that the items with the greatest utility are most likely to be selected. The selection will not always select the actual maximum option, but it can be shown that the mechanism performs optimally while still remaining differentially private. To use the exponential mechanism one evaluates the utility $\text{u}(\{\bm{y}_*,\bm{y}_t\},\theta_i)$ for a given database (which in our case consists of predictions and test outputs $\{\bm{y}_*,\bm{y}_t\}$) and for element $\theta_i$, from a set $\Theta$ (which consists of all the hyperparameter configurations we wish to compare). One also computes the sensitivity, $\Delta_u$, of this utility function. Then one selects an element $\theta_i$ with probability proportional to,

\begin{equation}
p_{\theta_i} \propto \exp \left( \frac{\varepsilon \text{u}(x,r)}{2 \Delta_u} \right).
\label{expmech}
\end{equation}

The rest of this section looks at a method for computing this utility and its sensitivity.

For the utility function we considered using the log marginal likelihood, with additional noise in the data-fit term to capture the effect of the DP noise. However for simplicity in estimating the sensitivity bound and to avoid overfitting we used the sum square error (SSE) over a series of $\kappa$-fold cross-validation runs, which for the $N_k$ points in a given fold is $\sum_{i=1}^{N_k} \left( y_{*i} - y_{\text{t}i} \right)^2$, where the two terms are elements from the predictions $\bm{y}_*$ and test values $\bm{y}_{\text{t}}$, respectively. This is somewhat similar to the method described in \citet{chaudhuri2011differentially}, but we need further thought regarding the sensitivity of the utility function. In \citet{chaudhuri2011differentially} this was simply equal to 1 as the utility was equal to the number of misclassifications. Also, by using the SSE the method will be applicable to other regression algorithms that do not provide a log likelihood.


Before proceeding we need to compute a bound on the sensitivity of the SSE. To briefly recap, the DP assumption is that one data point has been perturbed by at most $d$. We need to bound the effect of this perturbation on the SSE. First we realise that this data point will only be in the \emph{test set} in one of the $\kappa$ folds. In the remaining folds it will be in the training data.

If the perturbed data point is in the training data ($\bm{y}$), then we can compute the sensitivity of the SSE using the cloaking mechanism previously described. The perturbation this would cause to the predictions ($\bm{y}_*$) is described using standard GP regression (and the cloaking matrix). Specifically a change of $d$ in training point $j$ will cause a $d\bm{c}_{jk}$ change in the test point predictions, where $\bm{c}_{jk}$ is the $j$th column of the cloaking matrix for the $k$th fold.

To compute the perturbation in the SSE caused by the change in the training data, we note that the SSE is effectively the square of the Euclidean distance between the prediction and the test data. We are moving the prediction by $d\bm{c}_{jk}$. The largest effect that this movement of the prediction point could have on the distance between prediction and test locations is if it moves the prediction in the opposite direction to the test points. Thus it can increase (or decrease) the distance between the test and predictions by the largest length of $d\bm{c}_{jk}$ from all training points for fold $k$. Hence for one of the folds, the largest change in the SSE is $d^2 \max_j |\bm{c}_{jk}|_2^2$.

If the perturbed data point, $j$, was in the test data then the SSE will change by $\left( y_{*j} + d - y_{\text{t}j} \right)^2 - \left( y_{*j} - y_{\text{t}j} \right)^2 = d^2 + 2d (y_{*j} - y_{\text{t}j})$. The last part of the expression (the error in the prediction for point $j$) is unbounded. To allow us to constrain the sensitivity we enforce a completely arbitrary bound of being no larger than $\pm 4d$, thresholding the value if it exceeds this. Thus a bound on the effect of the perturbation if it is in the test set is $d^2 + 2d \times 4d = d^2 + 8d^2 = 9d^2$. The consequence of restricting this error is that we will underestimate the SSE when selecting a hyperparameter configuration, but this will only affect configurations with particularly high SSE, and thus will be those with very low probability anyway.

The SSE of each fold is added together to give an overall SSE for the cross-validation exercise. We sum the $\kappa-1$ largest training case sensitivities and add $9d^2$ to account for the effect of the single fold in which the perturbing data point, $j$, will be in the test set. The perturbation could have been in the test data in any of the folds. We assumed it was in the fold with the smallest training-data sensitivity to allow us to make the result a lower bound on the sensitivity of the SSE to the perturbation. If it had been in any other fold the sensitivity would have been less. Thus the sensitivity (for hyperparameter configuration $\theta_i \in \Theta$) of the SSE over the $\kappa$ folds is
\begin{equation}
\Delta_u^{(\theta_i)} = 9d^2 + \sum_{k=1}^{\kappa-1} d^2 \max_j |\bm{c}_{jk}|_2^2,
\end{equation} where the $\kappa$ folds are ordered by decreasing sensitivity (so that $k=1$ has the greatest, etc). The sensitivity for the exponential mechanism is the largest of all the possible configurations,
\begin{equation}
\Delta_u \triangleq \underset{\theta_i \in \Theta}{\max}\; {\Delta_u^{(\theta_i)}}.
\label{expsensitivity}
\end{equation} 

Thus, we compute the SSE and the SSE's sensitivity for each of the hyperparameter combinations we want to test and select the largest sensitivity from each combination to use as $\Delta_u$. We then use the computed sensitivity bound with the exponential mechanism to select the hyperparameters. So to reiterate, in this case each utility corresponds to a negative SSE, and the sensitivity, $\Delta_u$, corresponds to the the sum in \eqref{expsensitivity}. Equation \eqref{expmech} defines the probabilities we use to select the hyperparameters $\theta_i$ using the mechanism.


Note that for a given privacy budget, some $\varepsilon$ will need to be expended on this selection problem, and the rest expended on the actual regression.

A significant problem with the na\"{i}ve application of the above method is that the sensitivity, $\Delta_u$, is dominated by hyperparameter configurations which have the greatest sensitivity but are unlikely to have high utility. This leads to the probabilities of each configuration becoming almost equal. The sensitivity calculation depends on just $\bm{c}_{jk}$ and the specified data sensitivity, $d$. Thus it doesn't depend on the private outputs and therefore we can selectively ignore those configurations with a sensitivity above a certain threshold. This might seem to be potentially discarding an optimum configuration, however configurations in which the SSE is highly sensitive to a single training point are also those in which the predictions themselves are sensitive requiring destructive levels of DP noise. So it is likely they will not be the configurations that achieve the best SSE.

Finally, when estimating the SSE for each configuration we include the effect of the DP noise (specifically we sample many times from the DP noise distribution) to ensure that this is included in the final SSE. The noise is independent of private data so this addition doesn't introduce extra DP requirements. This is so that when selecting the configuration we take into account the cost of the DP noise. The full calculation is described in Algorithm \ref{alg}.

\begin{algorithm*}
  \newcommand*\Let[2]{\State #1 $\gets$ #2}
  \newcommand{\Model}{\mathcal{M}}
  \newcommand{\grad}{\frac{dL}{d\bm{\lambda}}}
  \newcommand{\lam}{\bm{\lambda}}
  \caption{Hyperparameter selection using the exponential mechanism.}
  \begin{algorithmic}[1]
    \Require{$\Model$ and $\Theta$; the GP model and the hyperparameter configurations we will test.}
    \Require{$X, \bm{y}$; training inputs and outputs}
    \Require{$X_{*} \in R^{P \times D}$, (the matrix of test inputs)}
    \Require{$d > 0$, data sensitivity (maximum change possible) \& $\varepsilon>0, \delta > 0$, the DP parameters}
    \Require{$\kappa > 1$, the number of folds in the cross-validation.}
    \Statex
    \Function{HyperparameterSelection}{$X$,$\bm{y}$,$X_{*}$, $M$, $\Theta$, $d$, $\varepsilon$, $\delta$, $\kappa$}
      \For {$\theta \in \Theta$}
        \Let{$\text{SSE}^{(\theta)}$}{0}
        \For {$k \in \kappa$}
          \Let{$C_k$}{$\Model\textsc{.get\_C}(X^{(k)},X_{*}^{(\bar{k})},\theta)$} \Comment{Compute the value of the cloaking matrix ($K_{*f} K^{-1}$) using $k$th fold's training data and testing on all the other data.}
          \Let{$\bm{y}_*^{(k)}$}{$C_k \bm{y}^{(k)}$}
          \Let{$\text{SSE}^{(\theta)}$}{$\text{SSE}^{(\theta)}+\sum_{i=1}^{N/\kappa} \left( y_{*i}^{(k)} - y_{\text{t}i}^{(k)} \right)^2$}
          \Let{$\alpha_k$}{$\max_j |\bm{c}_{jk}|_2^2$}
        \EndFor
        \State{Sort $\bm{\alpha}$ to be ascending}
        \Let{$\Delta_u^{(\theta)}$}{$9d^2 + \sum_{k=1}^{\kappa-1} d^2\alpha_k$}
      \EndFor
      \State{Select a hyperparameter configuration from $\Theta$ with probability proportional to,
      $\exp \left( \frac{\varepsilon \text{SSE}^{(\theta)}}{2 \Delta_u^{(\theta)}} \right)$}
    \EndFunction
    
  \end{algorithmic}
  \label{alg}
\end{algorithm*}

\section{Results}
We test the above methods with a series of experiments. Firstly, we investigate the nonstationary covariance methods, deployed to reduce outlier noise. We explored their effect in one and two dimensions in Sections \ref{kungdemo} and \ref{kungdemo2} respectively. In Section \ref{notesonvar} we look at reasons for why varying the lengthscale doesn't reduce sensitivity to outliers. We secondly look at the binary DP classification method. In Sections \ref{classification1} and \ref{classification2} we experiment in one and two dimensions. Thirdly, we test the combined sparse DP classification method on the high dimensional MNIST dataset. Finally, Section \ref{hyperparameter} demonstrates how DP (hyper)parameter selection might be performed.

\label{results}
\subsection{Non-stationary covariance: 1d !Kung}
\label{kungdemo}

In the examples below we experiment with both variable lengthscales and sparse approximations, as candidate methods for reducing the DP noise added in outlier regions.

\begin{figure}
  \centering
    \vspace{-3mm}  
    \includegraphics[width=1.0\columnwidth]{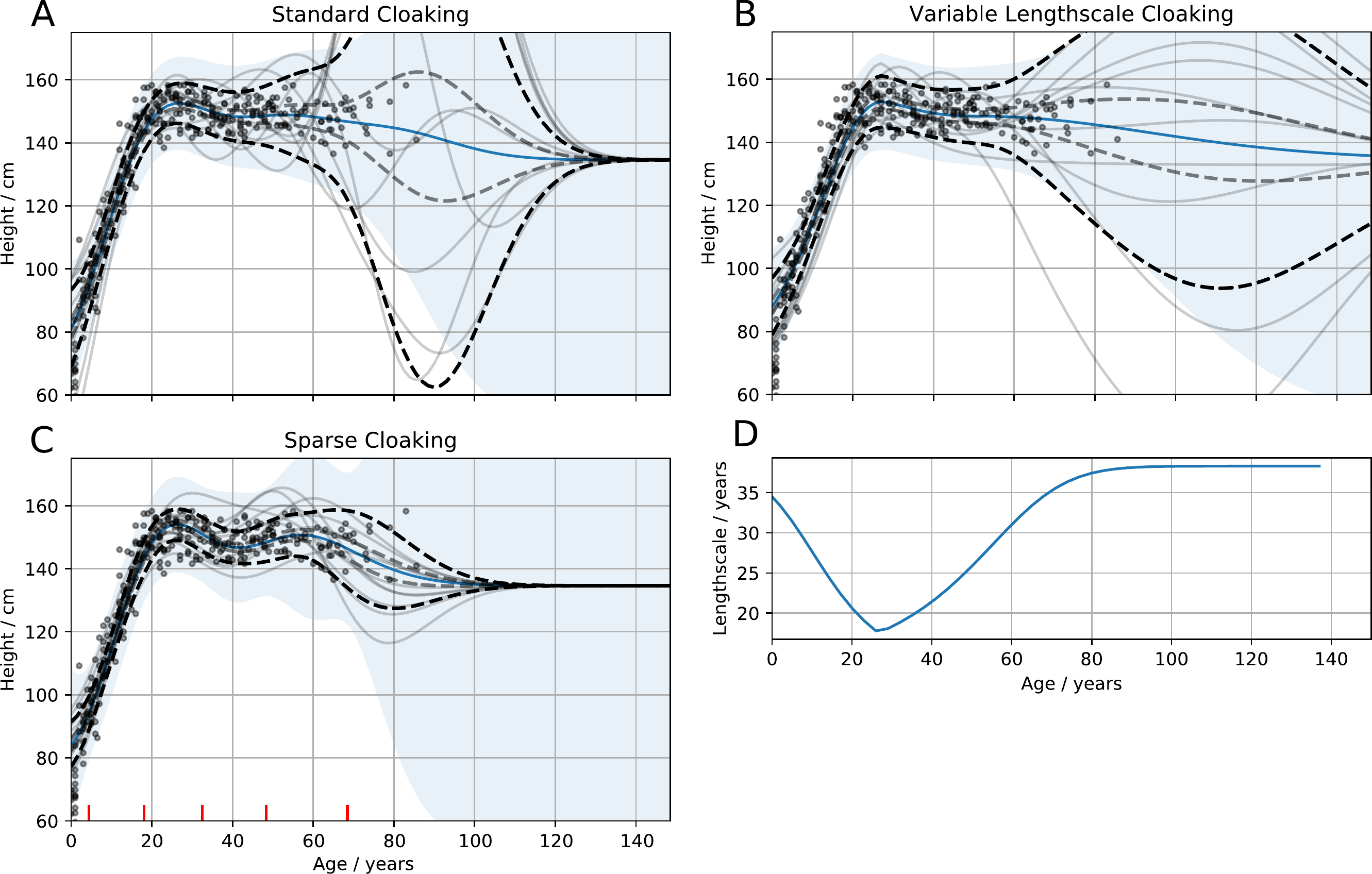}
  \caption{!Kung heights vs ages. In (A)-(C), solid blue lines, posterior means of the GPs; grey lines, DP samples; Black and grey dashed lines, standard error and $\frac{1}{4}$ standard error confidence intervals for DP noise respectively; blue area, GP posterior variance (excluding noise). $\varepsilon = 1$, $\delta=0.01$, $\Delta=100\;\text{cm}$. (A) Standard cloaking method. Considerable DP noise added in region of outliers (70-110 years). (B) Cloaking with variable lengthscale GP. (C) Cloaking, using five inducing inputs (placed using k-means clustering). (D) The varying lengthscale used in (B).}
    \vspace{-3mm}    
  \label{cloaking}
\end{figure}

Figure \ref{cloaking} demonstrates the three methods for reducing the DP noise added to the predictions. Figure \ref{cloaking}C shows the effect of the sparse cloaking method in which five inducing inputs, placed using k-means clustering, have been used to reduce the sensitivity to outliers. In figure \ref{cloaking}B, the variable lengthscale method is demonstrated. We found this method to perform poorly (see table \ref{tabledata}). Figure \ref{cloaking}D shows the lengthscale used. Specifically we've constructed the lengthscale function specifically to try and mitigate the poor results, by adjusting the lengthscale such that regions next to outlier areas also have relatively long lengthscales. The intention being that these dense `edge' regions can support predictions into the outlier areas. However we found we could only go so far, as increasing the lengthscale meant the GP would become unable to fit the true latent function's structure.

\subsection{Non-stationary covariance: 2d !Kung}
\label{kungdemo2}
\begin{figure}
\begin{center}
    \vspace{-15mm}
    \includegraphics[trim={4.5cm 0 3.7cm 0},width=0.4\columnwidth]{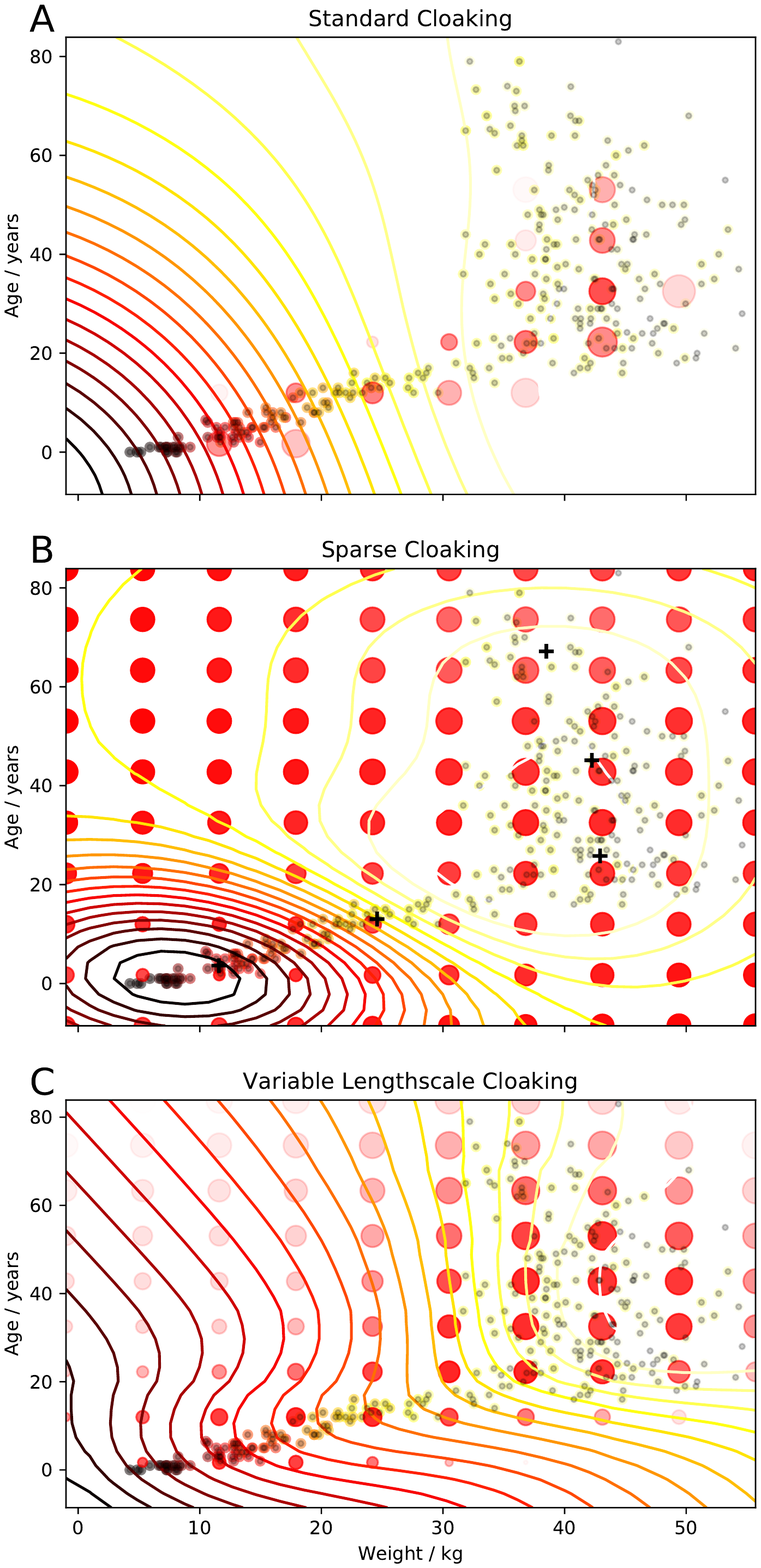}
    \includegraphics[trim={0cm -2cm 0cm 0},width=0.07\columnwidth]{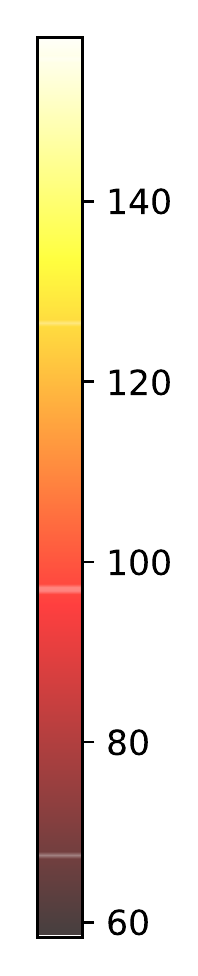}   
  \captionof{figure}{Predicting !Kung height from their weights and ages. Contours indicate non-DP posterior height predictions (in cm); dots, training inputs; disks indicate DP test predictions (with size proportional to predicted height; transparency to DP-noise, with full transparency reached when noise standard deviation reaches 40cm). $\varepsilon = 1$, $\delta=0.01$, $\Delta=100\;\text{cm}$. (A) Standard cloaking method. Only able to make low-noise predictions along part of a limited manifold (B) Sparse cloaking, using five inducing inputs (marked with crosses, placed using k-means clustering). DP noise reduced across whole domain. (C) Cloaking with variable lengthscale GP. Has slightly reduced DP noise added.}
  \label{cloaking_figures}
\end{center}
\end{figure}

At higher dimensions more of the data becomes an outlier (consider how there are more directions in which a data point can remove itself from the manifold in which the majority of the data lies). To demonstrate, we've added a new dimension (weight) to the !Kung data inputs (leaving the kernel as the exponentiated quadratic but now with two inputs). In figure \ref{cloaking_figures} the most opaque discs are those will the least DP noise. In the top figure, we see that the majority of the domain has so much DP-noise added that the discs have been rendered completely transparent, with only a tiny part of the input space available for making predictions. For example the noise added for a test point at 50kg, aged 60, has a standard deviation of 40cm (approximately 25\% of the non-DP prediction). Figures \ref{cloaking_figures}B and C illustrate the improvement in DP-noise when using the sparse method, or the variable lenthscale method.

Importantly in both non-stationary methods the DP-noise at the prediction is much reduced. However the non-DP mean predictions differ; with the inducing method inevitably having a far smoother surface due to the limited number of inducing inputs, and a relatively long lengthscale. In summary, much more of the domain can be used to make predictions without DP destroying the prediction.

To further illustrate, we performed a cross-validation run on the two datasets. One would expect that providing the additional information about their weight into the model, would improve the estimates of their height. However for the standard cloaking method, the additional DP noise caused by the increased sparsity in the data leads to an \emph{increase} in the RMSE, while the sparse cloaking method is not only better in the 1d case, but is not degraded by the additional weight information (and can use this to improve its predictions).

In summary, using a sparse approximation with inducing points placed at locations of high data density, we are able to perform DP GP regression over the whole domain without excessive noise in outlier regions. The alternative variable-lengthscale strategy however, doesn't appear to achieve the same goal.

\begin{table}
\begin{center}
\begin{tabular}{ c c c c }
  & standard cloaking & sparse & nonstationary \\ 
  \hline
  1d & $13.3 \pm 3.2$ & $9.9 \pm 1.7$ & $12.8 \pm 2.0$ \\
  2d & $17.2 \pm 3.1$ & $10.2 \pm 1.6$ & $16.3 \pm 2.5$ \\
  1d (no DP) & $7.1 \pm 1.8$ & $7.7 \pm 1.9$ & $8.6\pm2.2$ \\
  2d (no DP) & $5.2 \pm 1.5$ & $7.7 \pm 2.0$ & $5.9 \pm 1.7$ \\
\end{tabular}
\end{center}
\caption{RMSEs estimating height (cm) from weight and age. Using 14-fold cross-validation runs on the !Kung dataset. First two rows use $\varepsilon=1.0$, $\delta=0.01$, $\Delta=100\text{cm}$. The second two rows have no DP noise added. Standard deviation over 14-folds indicated by $\pm$.\label{tabledata}}
\end{table}


\subsection{Notes on the variable lengthscale results}
\label{notesonvar}
We briefly look at why the non-stationary kernel solution is not achieving the desired reduction in DP noise. Recall that when making predictions we want to average over sufficient training points so that individual training points only have a small effect on the prediction. Thus we want, for a test point in the region of outliers, to still have a high covariance with many training points necessarily including some from high density regions. Conversely we want a test point in the middle of the high density region to only have a high covariance with nearby training data (i.e. short lengthscale) to allow it to describe more complex structure. 

We can see, with a little thought, that these two  requirements can't both be met. Consider a pair of points, A and B, within the short-lengthscale regime and a third point, C, in the long-lengthscale regime such that the three points are evenly spaced. We want, for predictions near C, a high covariance between the pairs A-C and B-C, but we also want a low covariance between the short-lengthscale pair A-B. We might also want a lower-covariance between A-C when predicting near A - the opposite of that required for the predictions at C.

It is informative to consider how Gibbs developed the kernel. He first considers an exponentiated quadratic (EQ) $L$-dimensional basis function centred at $\bm{a}_k$ with parameterised radius $r_l(\bm{x};\Theta_l)$,
$$
\phi_k(\bm{x}) = \prod_{l=1}^L \sqrt{\frac{\sqrt{2}}{r_l(\bm{x};\Theta_l)}} 
\exp \Bigg[-\sum_{l=1}^L \frac{(x^{(l)}-a_k^{(l)})^2}{r_l^2(\bm{x};\Theta_l)} \Bigg].
$$
The prefactor here ensures that the resulting covariance function has a constant variance (wrt $\bm{x}$). Gibbs goes on to build the covariance by considering the integral over the product of such kernels, resulting in the covariance,
\begin{multline}
C(\bm{x}_m,\bm{x}_n; \Theta) = \theta_1 \prod_l\Bigg \{\frac{2 r_l(\bm{x}_m; \Theta) r_l(\bm{x}_n; \Theta)}{r_l^2(\bm{x}_m;\Theta) + r_l^2(\bm{x}_n;\Theta)} \Bigg \}^{1/2}\\
\times \exp \Bigg(-\sum_l{\frac{\left(x_m^{(l)}-x_n^{(l)}\right)^2}{r_l^2(\bm{x}_m;\Theta)+r_l^2(\bm{x}_n;\Theta)}} \Bigg).
\label{covariance_equation}
\end{multline}
Considering just the integral over the product of a pair of bases. If two bases have long lengthscales, then even if the inputs are far apart this product will be large, as they have similar distributions that mostly overlap. If the bases have short lengthscales however, they will not overlap and the covariance will thus be small. Of greater interest for our problem is the mixed case, in which the pair of inputs are in long- and short-lengthscale regions. Even if the two inputs are close together the two distributions are poorly matched and thus their product will be small, leading to a low covariance. Specifically if we consider two nearby points (separated by distance $d$), we differentiate \eqref{covariance_equation} with respect to the lengthscales of input $m$, $l_m = r_l(\bm{x}_m;\Theta)$, and set to zero. The choice of $l_m$ which will maximise the covariance approaches the other input's covariance $l_n = r_l(\bm{x}_m;\Theta)$ in the limit as $l_n >> d$. Simply put, locations with wildly different lengthscales will have a low covariance.

As an alternative we could considered a weighted sum of GPs as proposed by \citet{herlands2016scalable} in which the proportion of each function varies over the domain. We found this suffers from the same restriction. Briefly, if we simplify equations (8) and (9) from \citet{herlands2016scalable} thus;

$$
y(x) = w(x)f(x) + (1-w(x))g(x),
$$ 

where $w$ is a function describing the weighted summing of the two GP functions $f$ and $g$. The covariance of this new function is;

$$k(x,x') = w(x)k_f(x,x')w(x') + \\(1-w(x))k_g(x,x')(1-w(x'))$$

If the two points $x$ and $x'$ are within the same regime, then $w(x)$ and $w(x')$ will be roughly equal. However if the two points are from different regions (e.g. if $x$ were in a long-lengthscale region, while $x'$ is associated with a short lengthscale) the products $w(x)w(x')$ or $(1-w(x))(1-w(x'))$ will be very small, leading to a small covariance between points across the two regimes, even if those points are nearby.

In summary predictions in the outlier regions will not be supported by training points within the short-lengthscale domain.

\subsection{Classification: Defaulting example}
\label{classification1}
To illustrate the classification method, we use the Home Equity Loans (HEL) dataset \citep{scheule2017credit}. We start with a one-dimensional example, using as an input the number of delinquent credit lines (DELINQ). Realistically this is likely to be a feature that an individual would prefer to remain private. We will however leave the provision of input privacy for future work. Instead we assume that it is just the outputs (whether they have defaulted, BAD) that we wish to keep private.

\begin{figure}
  \centering
    \vspace{-3mm}  
    \includegraphics[width=0.7\columnwidth]{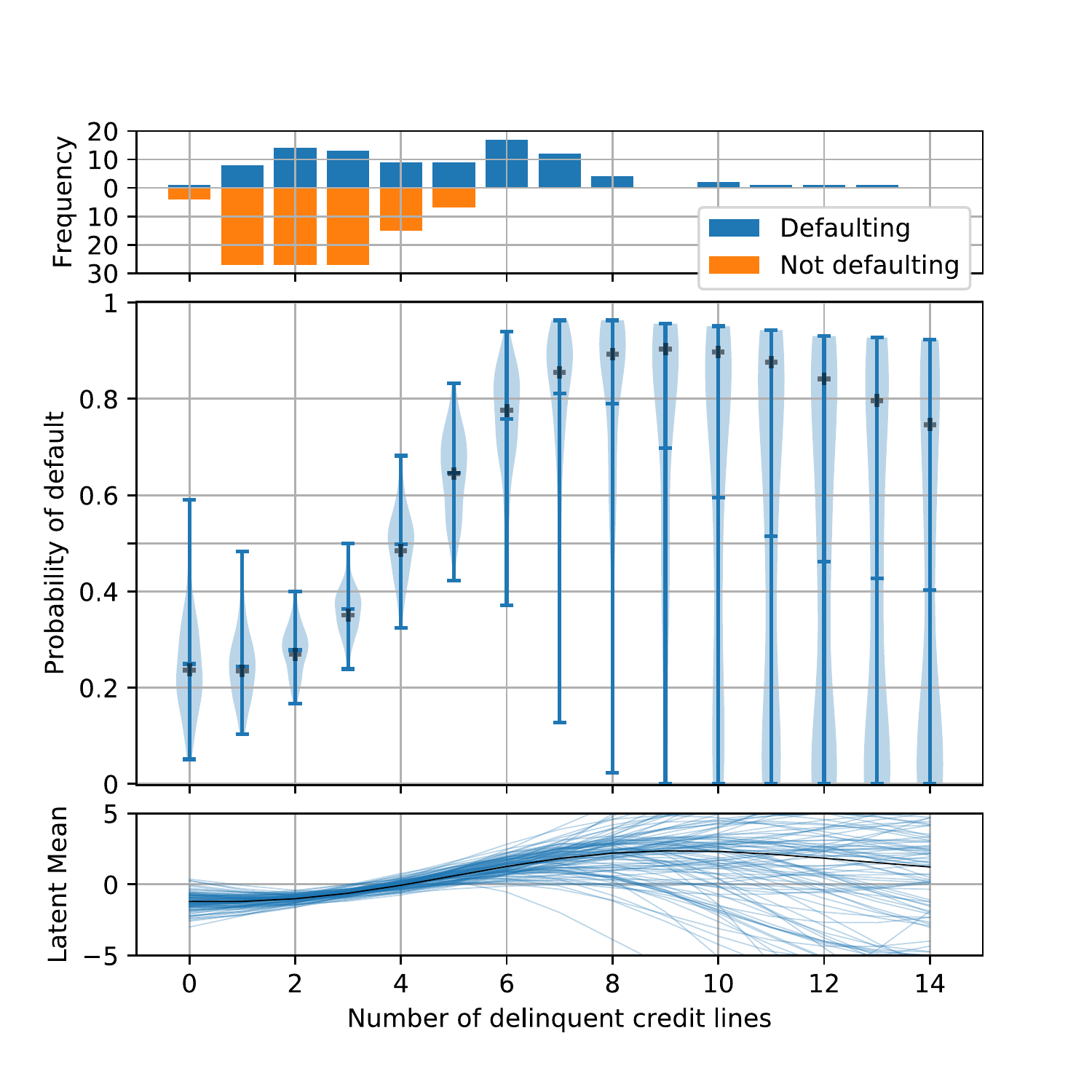}
  \caption{Home Equity Loans dataset classification example. A single dimension (number of delinquent credit lines, DELINQ) is used to predict the likelihood of a borrower defaulting on a loan. We assume for the purpose of this demonstration that the training inputs, DELINQ are not private, but the default status is. The upper plot shows the frequency of defaulting for each delinquency count, the middle plot shows the distribution of one hundred DP samples. The black crosses indicate the non-DP mean. The lower plot shows the associated DP latent means (non-DP mean in black). \label{hel_classification1d}}
    \vspace{-3mm}
  
\end{figure}

We bias sample two hundred points from the HEL dataset to approximately balance both the DELINQ input and the BAD outputs. The upper part of figure \ref{hel_classification1d} indicates the frequencies for each category. The middle part of the figure shows, using a violin plot, the DP predicted probabilities associated with each possible delinquency count between zero and fourteen. The black crosses indicate the non-DP predicted probabilities. At low counts the risk of default is low but increases with greater numbers of delinquent credit lines. Returning to the distribution of DP posterior means, as the amount of data decreases (over five or six credit lines) we begin to see divergence between the non-DP mean (black crosses) and the means produced with DP noise added during estimation. Specifically, the actual mean remains relatively confident (above 70\%), but the DP mean (with noise added) becomes spread across almost the whole range, with concentrations at the extremes due to the squashing effect of the link function.

\subsection{Classification: 2d example}
\label{classification2}
\begin{figure}
  \centering
    \vspace{-3mm}  
    \includegraphics[width=0.7\columnwidth,trim={2.8cm 2.8cm 2.8cm 2.8cm},clip]{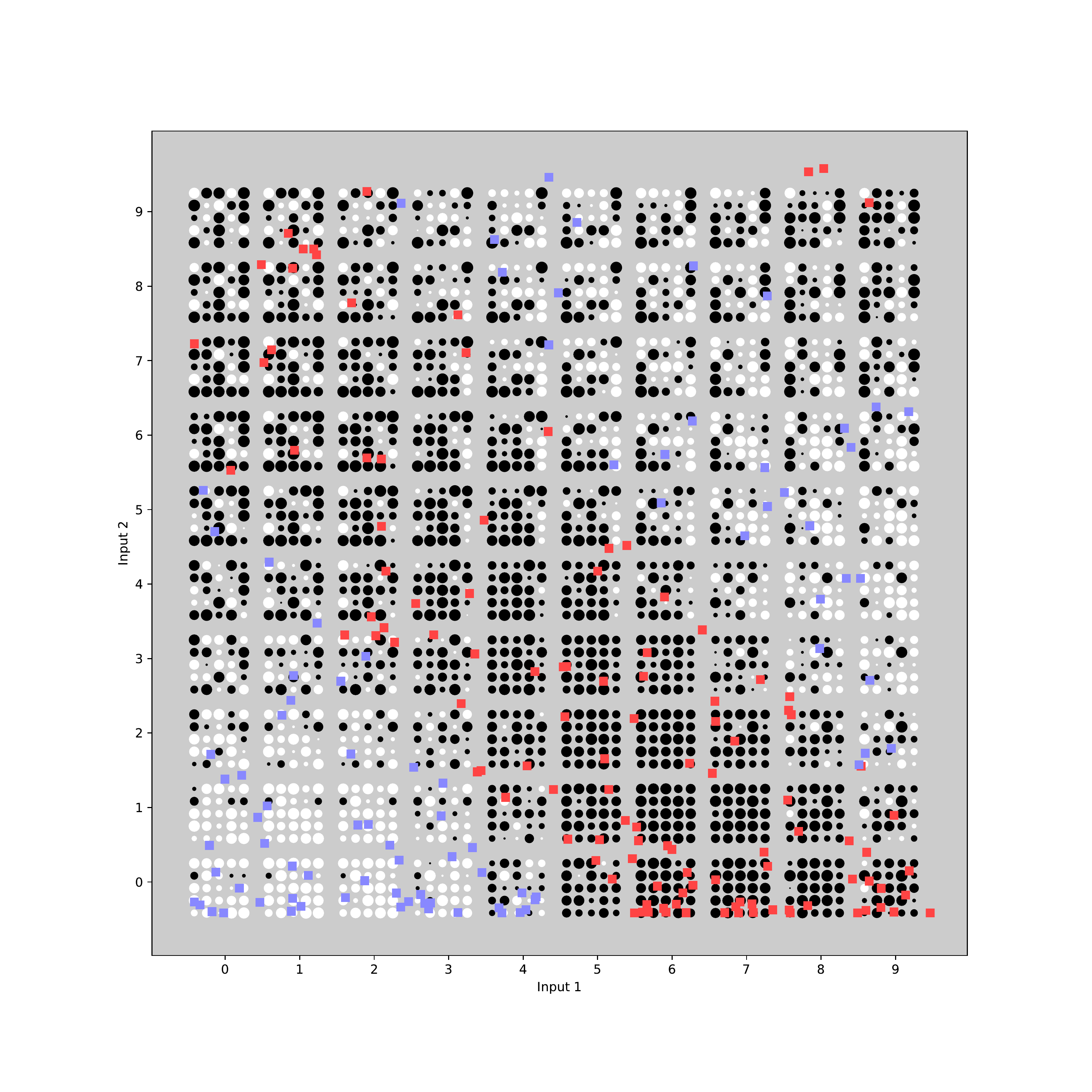}
  \caption{Grouped Hinton plot showing the posterior mean for 25 DP-classification attempts. Each block of circles describes 25 different DP predictions of the transformed posterior mean at each integer location. The white and black circles correspond to values over and under 0.5, respectively. The size of the circle expresses confidence and indicates the distance from 0.5, so the largest circles are either 0 (black) or 1 (white). The same DP sample is at the same relative location in each block. The training inputs are simulated two dimensional values with the blue (+ve) and red (-ve) crosses indicating their location and class. $l_1=l_2=3.5$. 200 iterations were computed for the cloaking algorithm. 1 iteration of the $\hat{f}$ optimisation was performed. $\varepsilon = 1.0$, $\delta = 0.01$. Sensitivity = 2 (-1 to +1). \label{sim_classification2d}}
    \vspace{-3mm}
  
\end{figure}

To very briefly demonstrate the classification algorithm operating in higher dimensions and to illustrate a slightly more convoluted function, we simulated a simple two-dimensional training set of 200 points consisting of a series of noisy diagonal binary class stripes (figure \ref{sim_classification2d}). The figure also shows predictions from 25 separate DP-classification computations. A concentration gradient has been included in the simulated training data. Near the bottom of the image the concentration is high and the DP samples have little variation. Near the training data's class boundaries the predictions are nearer 0.5 (less confident) indicated by smaller dots (e.g. at location [3,2]). Towards the top of the plot (and to an extent around the edge) the training data becomes sparse, leading to greater DP noise manifesting itself as a mix of highly confident predictions for both classes. The scale of the noise could, in future, be included in the test point uncertainty reported.

The method used runs the optimisation step (finding $\hat{f}$) of the Laplace approximation only once. We briefly tried dividing the privacy budget between two iterations, but found the result was considerably worse. We used a test set spaced over the $10 \times 10$ grid (consisting of 50 of each class) and computed 25 different DP noise results. We found the accuracy fell from 69\% to 51\% with the second iteration, suggesting that the additional noise required to ensure privacy in two steps of optimisation far outweighs the potential benefits of an improved estimate of $\hat{f}$. Note the accuracy without DP noise was 81\% (with 20 iterations of the optimiser), obviously better classifiers exist, and the choice of hyperparameters was deliberately chosen to aid the DP prediction potentially at the cost of the non-DP prediction's accuracy.

\subsection{Sparse Classification: MNIST}
\label{mnist}
\begin{figure}
  \centering
    \vspace{-3mm}  
    \includegraphics[width=0.7\columnwidth]{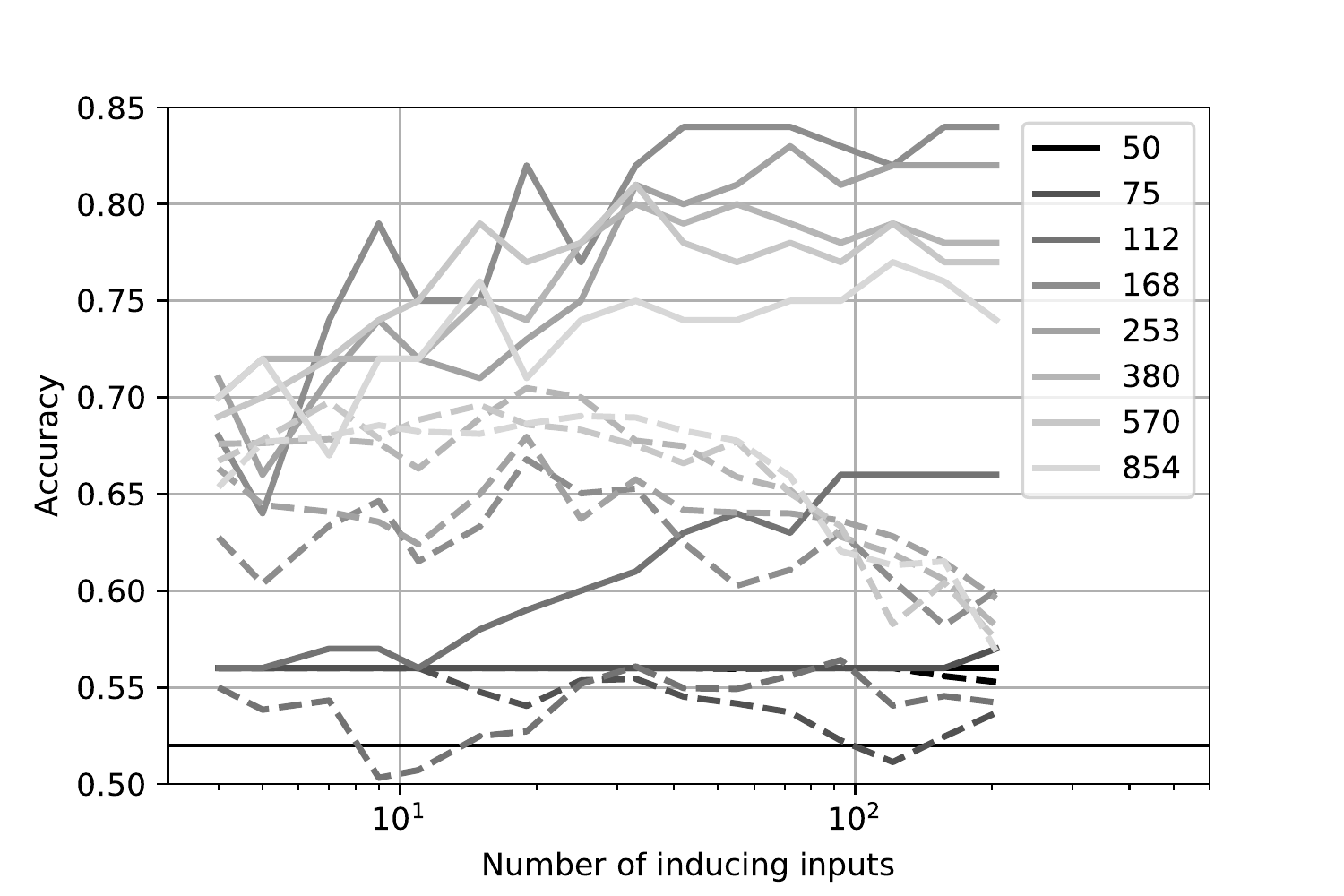}
  \caption{Graph showing accuracy of sparse GP classifier with DP (dashed lines) or without DP (solid lines). Chance accuracy at 52\% indicated by horizontal black line. x-axis indicates number of inducing inputs (between 4 and 204, N=256). Shade indicates lengthscale. Pixel values in the data varied from 0 to 255, so the lengthscale spans this, with black being short lengthscales and light grey, long. For the non-DP classification there is an optimal lengthscale of about 168, and as one might expect, the accuracy improves with more inducing inputs. The DP method on the other hand is optimal with longer lengthscales, presumably because this allows more training points to be averaged. It also peaks at lower numbers of inducing inputs. As discussed for the sparse regression results, this is due to reducing outlier influences. DP points computed from 25 DP model optimisation runs leading to a maximum SE of 1.86\% accuracy. Associated CIs omitted for clarity. The number of inducing inputs was tested at sixteen exponentially spaced values (4 to 200). \label{clusters}}
    \vspace{-3mm}
  
\end{figure}

To demonstrate the DP sparse classification method we turn to a higher dimensional dataset. In this problem we try to classify MNIST digits as being in either the low class or high class, \{0,1,2,3,4\} or \{5,6,7,8,9\} respectively. The MNIST data points were down-sampled to $15 \times 15$ images. We trained the algorithm with 256 points and tested on 100 other points. Figure \ref{clusters} demonstrates the impact of both the number of inducing inputs and the lengthscale. The inducing inputs were again chosen using k-means clustering, The lengthscales were all equal and the kernel variance fixed at one. We found that unlike the non-DP results, the DP method favoured longer lengthscales and fewer inducing inputs. The former result probably due to the averaging effect longer lengthscales achieve (each individual training point has less influence). The latter result, with fewer inducing inputs providing better predictions is due to the down-weighting of outliers and instead reflects the behaviour of the output around where the data was concentrated. It potentially appears to peak around 21 inducing inputs, and has a marked decline above 40 inducing inputs. We conclude that the inducing inputs help reduce the spurious DP noise one would normally have to introduce to protect the outliers as we found in the regression example in Section \ref{kungdemo2}. This dataset sits on a relatively low-dimensional manifold due to the correlations between neighbouring pixels and the relative similarity of training data points, hence it was concentrated in such a way that only a few inducing inputs were able to approximate it quite well, not necessarily a feature of all high-dimensional datasets.

\subsection{!Kung Hyperparameter Selection}
\label{hyperparameter}
For our 1d !Kung example our model has three parameters, the lengthscale, the kernel variance and the likelihood's Gaussian noise variance. Note that for simplicity we used the standard cloaking method without sparse approximation. We need to select these in such a way that the training data remains private while the predictions have the lowest SSE possible. For this experiment we split the data into two halves; using one half for selecting hyperparameters, and the other for estimating the SSE achieved if we had used those hyperparameters. The training half is again split in a 5-fold cross-validation step to estimate the SSE for each configuration of hyperparameters. This leaves only 115 points for training. The sensitivity for the SSE is computed based on this cross-validation. We can then report the expected SSE of the training set using the probabilities provided by the exponential mechanism.

In this experiment we search exponentially the three parameters (lengthscale 1, 5, 25, 125, 625; Gaussian noise variance 0.2, 1, 5, 25; Kernel variance 1, 5, 25, 125). The full table of RMSEs and probabilities is in the supplementary. The configuration that achieved the lowest error in the training set was found to be with lengthscale, 25; Gaussian noise variance, 25; kernel variance, 1.0; with a RMSE of 14.55cm. The long lengthscale is particularly relevant here as, unlike in Section \ref{kungdemo}, we only have 115 training points, so averaging over the data provides the best compromise between an accurate fit and avoiding excessive DP noise. If we were to simply select this configuration we would be leaking private data through the computation of the SSEs. To avoid this we select a configuration at random with probability calculated using the exponential mechanism ($\varepsilon=1$). Although the above configuration had the least training RMSE it will still only selected with a probability of 0.0244, but this is almost twice the 1/80 average (given we tested 80 configurations). The expected RMSE (from the weighted sum of the RMSEs of the available configurations) is 19.02cm. While the mean of the 80 configurations is 153.81cm. Clearly picking a configuration using the exponential mechanism is better than randomly choosing, but not as good as the best configuration. We used up $\varepsilon=2$ DP, overall ($\varepsilon=1$ for the exponential mechanism and $\varepsilon=1$ for the DP GP regression). The trade-off in how this budget is left for future experimentation.

\begin{figure}[t!]
  \centering
    \vspace{-3mm}
    \includegraphics[width=0.5\columnwidth]{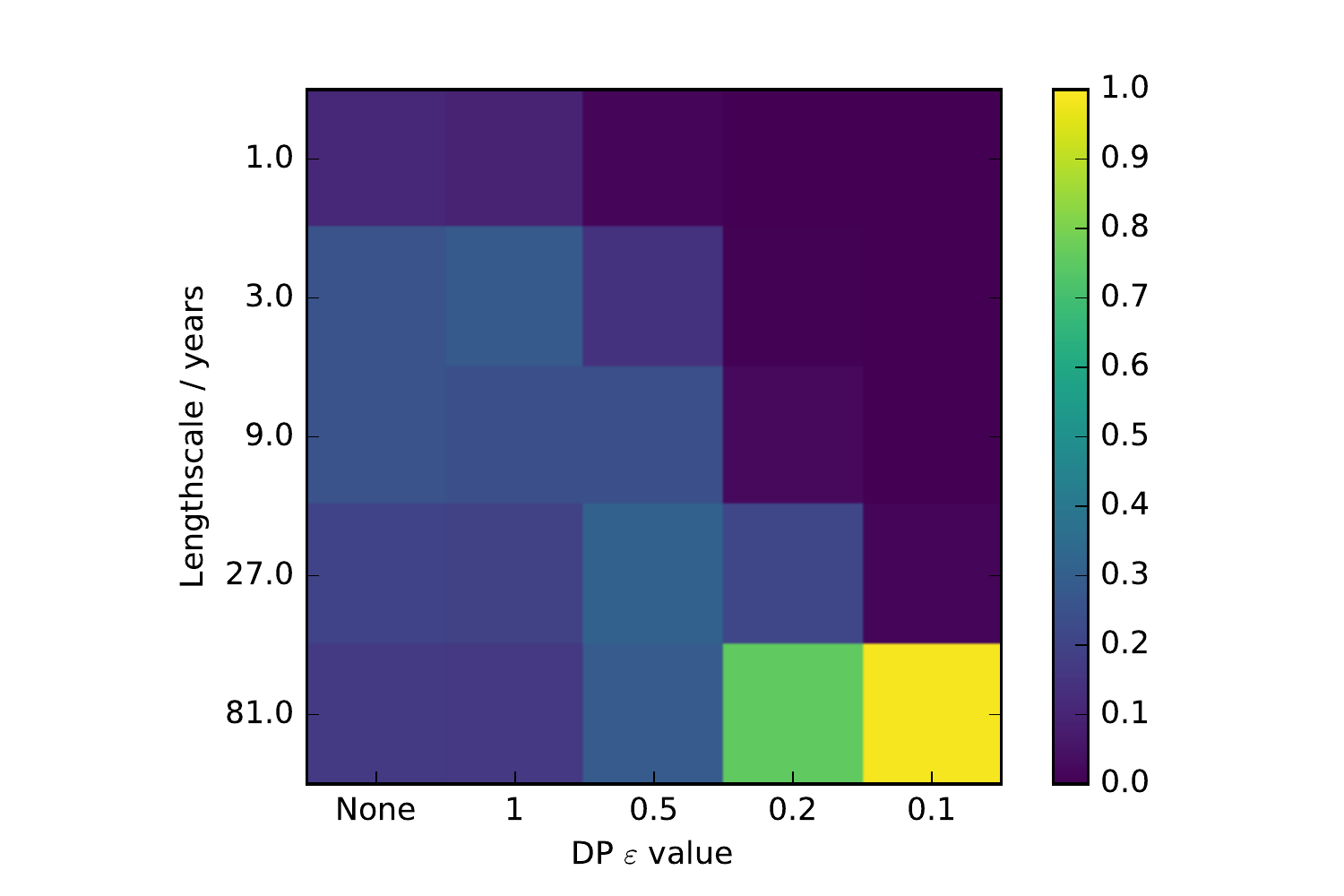}
    \vspace{-3mm}
  \caption{Effect of varying the differential privacy parameter, $\varepsilon$, on the likelihood of selecting each lengthscale. Colour indicates probability of parameter selection. With low privacy, a short lengthscale is appropriate which allows the GP to describe details in the data. With high privacy, a longer lengthscale is required, which will average over large numbers of individual data points. Used the whole dataset with 5-fold cross-validation.}
  \vspace{-3mm}
  \label{effectofepsonls}
\end{figure}

An interesting result is in the effect of the level of privacy in the regression stage on the selection of the lengthscale. This is demonstrated in the distribution of probabilities over the lengthscales when we adjust $\varepsilon$. Figure \ref{effectofepsonls} demonstrates this effect using the !Kung example. Each column is for a different level of privacy (from none to high) and each tile shows the probability of selecting that lengthscale. For low privacy, short lengthscales are acceptable, but as the privacy increases, averaging over more data by increasing the lengthscale allows us to mitigate the increasing DP noise.

\section{Discussion}
\label{discussion}
We have investigated several ways of expanding the range and power of DP Gaussian process models. We first looked at methods for reducing the excessive noise added in outlier regions of the dataset, an issue we showed would particularly need addressing if one wanted to do inference at increased dimensionality. Specifically we considered both sparse approximations and methods to manipulate the lengthscale. We found that the latter failed to achieve the expected gains as the data within the short-lengthscale high-density regions was largely uncorrelated with the outliers, leaving them with little support. However the sparse method, using inducing inputs, was able to substantially reduce the DP noise without compromising the DP guarantees.
Both sparse and nonstationary methods require additional parameters to be selected. Specifically, the sparse method requires the number and placement of inducing inputs to be chosen, while the variable lengthscale method requires a lengthscale function to be defined. In this case the inducing inputs were placed using simple k-means clustering. The lengthscale function required more attention and care. Note that in practice selecting a function with awareness of the data will lead to unintentional privacy-loss. Thus, from a practical, privacy perspective the sparse, inducing-input method appears to offer a safer easier to apply solution, which also performs substantially better. We applied a similar sparse-approximation to the problem of classification, for a larger scale dataset and found it also improved the results if we reduced the number of inducing inputs. It also confirmed the DP method could function with high-dimensional data, although the structure of the points within the domain may be important. We hypothesise that the method is most effective when the data lies on a relatively low-dimensional manifold, or concentrated into relatively few locations in the input domain.

A brief note on the placement of the inducing inputs. We could use a more traditional method that minimises a KL divergence to find the optimum locations. However this requires `peeking' at the outputs. Thus we would need to use a gradient descent method that is privacy-aware \citep{abadi2016deep,song2013stochastic} and bound the sensitivity of the training data on these gradients. More fundamentally, placement using such a standard sparse approximation method would place the inducing inputs in potentially very suboptimal locations in the DP situation. For example, a single outlier often is given its own inducing input when trying to minimise the KL divergence (if there's sufficient inducing inputs available) which will lead to massive DP noise added in the vicinity of this outlier. 

We next investigated DP GP classification. We made the predictions DP in the output by altering the internal iterative update that finds the mode of the posterior. It was found however that updating this more than once reduces the classification accuracy, as the additional DP noise required to span two updates outweighs the possible benefit from a better estimate of the mode. This might change with more training data, or additional dimensions. The latent function's DP mean behaves as in the regression case and contains more variance in outlier regions, even though the mechanism for the application of DP is somewhat different. The unexpected utility of the cloaking mechanism to this problem suggests that the cloaking method may have other applications, in which a vector of training data is linearly transformed. In future work the Gaussian noise of the DP should be included in the computation of the posterior (through the squashing, logistic function) rather than just use the variance of the latent function, as this will allow this component of the uncertainty to be included in any prediction. We excluded it from the posterior for clarity in the plots and results (allowing the DP and GP uncertainty to be distinguished).

We finally look at the issue of hyperparameter optimisation. We found that a crude grid search combined with the DP exponential mechanism provides a simple but effective method for selecting hyperparameters. Its efficacy may be down to the relatively low sensitivity of the RMSE to parameter selection, with many configurations resulting in similar error rates. The table in the supplementary material demonstrates this, with nearly half the configurations having a RMSE lying within 30\% of the best.

In summary, we have presented three tools to extend the use of differential privacy for Gaussian process models. First, the sparse approximation works by effectively reducing the effect of the outliers, allowing the model to make predictions in both dense and outlier regions. This becomes particularly important at higher dimensions. Second we devised a method for making a GP classification model differentially private, by altering its internal optimisation step. We found it was able to make predictions with a useful accuracy and with a relatively small number of training points. Finally we suggest a method for hyperparameter optimisation based on a straightforward grid search combined with the differentially private exponential mechanism.


\acks{This work has been supported by the Engineering and Physical Research Council (EPSRC) Research Project EP/N014162/1. We also thank Wil Ward for his assistance and suggestions.}

\newpage

\appendix
\section*{Appendix A.}

\begin{table*}
\footnotesize
\begin{minipage}[b][7cm][b]{.55\linewidth}
\hspace{-0.8cm}
\begin{tabular}{ r r r r r r }
LS & Noise var & Kernel var & Prob & RMSE & \\ 
\hline
1.0 & 0.2 & 1.0 & 0.0000 & 235.52 & \\
1.0 & 0.2 & 5.0 & 0.0000 & 436.54 & \\
1.0 & 0.2 & 25.0 & 0.0000 & 526.11 & \\
1.0 & 0.2 & 125.0 & 0.0000 & 715.99 & \\
1.0 & 1.0 & 1.0 & 0.0000 & 132.11 & \\
1.0 & 1.0 & 5.0 & 0.0000 & 244.78 & \\
1.0 & 1.0 & 25.0 & 0.0000 & 305.81 & \\
1.0 & 1.0 & 125.0 & 0.0000 & 430.29 & \\
1.0 & 5.0 & 1.0 & 0.0000 & 73.59 & \\
1.0 & 5.0 & 5.0 & 0.0000 & 162.80 & \\
1.0 & 5.0 & 25.0 & 0.0000 & 219.82 & \\
1.0 & 5.0 & 125.0 & 0.0000 & 283.01 & \\
1.0 & 25.0 & 1.0 & 0.0190 & 25.61 & \\
1.0 & 25.0 & 5.0 & 0.0000 & 60.73 & \\
1.0 & 25.0 & 25.0 & 0.0000 & 215.35 & \\
1.0 & 25.0 & 125.0 & 0.0000 & 248.26 & \\
5.0 & 0.2 & 1.0 & 0.0000 & 75.13 & \\
5.0 & 0.2 & 5.0 & 0.0000 & 94.60 & \\
5.0 & 0.2 & 25.0 & 0.0000 & 128.16 & \\
5.0 & 0.2 & 125.0 & 0.0000 & 233.28 &\\
5.0 & 1.0 & 1.0 & 0.0000 & 52.59 &\\ 
5.0 & 1.0 & 5.0 & 0.0000 & 80.72 & \\
5.0 & 1.0 & 25.0 & 0.0000 & 95.11 & \\
5.0 & 1.0 & 125.0 & 0.0000 & 167.44 &\\
5.0 & 5.0 & 1.0 & 0.0131 & 33.87 & \\
5.0 & 5.0 & 5.0 & 0.0000 & 133.09 &\\
5.0 & 5.0 & 25.0 & 0.0000 & 81.38 &\\
5.0 & 5.0 & 125.0 & 0.0000 & 187.44 &\\
5.0 & 25.0 & 1.0 & 0.0222 & 18.27 &\\
5.0 & 25.0 & 5.0 & 0.0122 & 34.01 &\\
5.0 & 25.0 & 25.0 & 0.0000 & 53.16 &\\
5.0 & 25.0 & 125.0 & 0.0000 & 80.85 &\\
25.0 & 0.2 & 1.0 & 0.0151 & 25.04 &\\
25.0 & 0.2 & 5.0 & 0.0000 & 32.92 &\\
25.0 & 0.2 & 25.0 & 0.0000 & 60.83 &\\
25.0 & 0.2 & 125.0 & 0.0000 & 43.17 &\\
25.0 & 1.0 & 1.0 & 0.0200 & 23.92 &\\
25.0 & 1.0 & 5.0 & 0.0181 & 26.28 &\\
25.0 & 1.0 & 25.0 & 0.0000 & 32.66 &\\
25.0 & 1.0 & 125.0 & 0.0000 & 42.26 &\\
\end{tabular}
\end{minipage}
\begin{minipage}[b][1cm][b]{.45\linewidth}
\begin{tabular}{ r r r r r r }	

LS & Noise var & Kernel var & Prob & RMSE & \\ 
\hline
25.0 & 5.0 & 1.0 & 0.0231 & 16.36 &\\
25.0 & 5.0 & 5.0 & 0.0210 & 20.89 &\\
25.0 & 5.0 & 25.0 & 0.0177 & 26.84 &\\
25.0 & 5.0 & 125.0 & 0.0000 & 31.85 &\\
25.0 & 25.0 & 1.0 & 0.0244 & 14.55 & ***\\
25.0 & 25.0 & 5.0 & 0.0233 & 16.13 &\\
25.0 & 25.0 & 25.0 & 0.0205 & 20.81 &\\
25.0 & 25.0 & 125.0 & 0.0180 & 27.74 &\\
125.0 & 0.2 & 1.0 & 0.0233 & 16.53 &\\
125.0 & 0.2 & 5.0 & 0.0223 & 18.02 &\\
125.0 & 0.2 & 25.0 & 0.0219 & 19.27 &\\
125.0 & 0.2 & 125.0 & 0.0211 & 20.90 &\\
125.0 & 1.0 & 1.0 & 0.0236 & 16.07 &\\
125.0 & 1.0 & 5.0 & 0.0233 & 16.29 &\\
125.0 & 1.0 & 25.0 & 0.0222 & 18.24 &\\
125.0 & 1.0 & 125.0 & 0.0225 & 18.78 &\\
125.0 & 5.0 & 1.0 & 0.0226 & 17.43 &\\
125.0 & 5.0 & 5.0 & 0.0236 & 16.00 &\\
125.0 & 5.0 & 25.0 & 0.0230 & 16.72 &\\
125.0 & 5.0 & 125.0 & 0.0228 & 17.62 &\\
125.0 & 25.0 & 1.0 & 0.0225 & 18.10 &\\
125.0 & 25.0 & 5.0 & 0.0227 & 17.49 &\\
125.0 & 25.0 & 25.0 & 0.0235 & 16.06 &\\
125.0 & 25.0 & 125.0 & 0.0232 & 16.65 &\\
625.0 & 0.2 & 1.0 & 0.0230 & 16.83 &\\
625.0 & 0.2 & 5.0 & 0.0242 & 14.72 & **\\
625.0 & 0.2 & 25.0 & 0.0232 & 17.06 &\\
625.0 & 0.2 & 125.0 & 0.0225 & 17.75 &\\
625.0 & 1.0 & 1.0 & 0.0224 & 18.09 &\\
625.0 & 1.0 & 5.0 & 0.0231 & 16.90 &\\
625.0 & 1.0 & 25.0 & 0.0242 & 14.61 & **\\
625.0 & 1.0 & 125.0 & 0.0231 & 16.59 &\\
625.0 & 5.0 & 1.0 & 0.0224 & 18.14 &\\
625.0 & 5.0 & 5.0 & 0.0225 & 18.02 &\\
625.0 & 5.0 & 25.0 & 0.0230 & 16.83 &\\
625.0 & 5.0 & 125.0 & 0.0241 & 14.52 & *\\
625.0 & 25.0 & 1.0 & 0.0226 & 17.71 &\\
625.0 & 25.0 & 5.0 & 0.0223 & 18.33 &\\
625.0 & 25.0 & 25.0 & 0.0224 & 17.88 &\\
625.0 & 25.0 & 125.0 & 0.0231 & 16.82 &\\
\end{tabular}
\end{minipage}
\caption{Hyperparameter search using the exponential mechanism. The top four parameter configurations that are most likely to be chosen are marked with asterisks. The probability is based on the SSE from the training fold while the RMSE is from the test fold. Many probabilities are exactly zero due to the thresholding of the SSEs.}
\label{hyperselection}
\end{table*}

Table \ref{hyperselection} reports the probability and RMSE for the combinations used in the hyperparameter optimisation exercise in section \ref{hyperparameter}.


\vskip 0.2in
\bibliography{refs}

\end{document}